\documentclass{article}
\usepackage{ijcai18}

\usepackage{amsmath}
\usepackage{amssymb}
\usepackage{times}  
\usepackage{helvet}  
\usepackage{courier}  
\usepackage{url}  
\usepackage{graphicx}  
\usepackage{xcolor}
\usepackage[vlined,ruled,linesnumbered]{algorithm2e}

\usepackage{xcolor}
\usepackage{soul}
\usepackage[utf8]{inputenc}
\usepackage[small]{caption}

\title{Dynamically Hierarchy Revolution: DirNet for Compressing Recurrent Neural Network on Mobile Devices}
\newcommand{\pname}{DirNet}
\author{Jie Zhang$^1$\thanks{Work was done during the first author's internship at Samsung Research America.}, Xiaolong Wang$^2$\thanks{Corresponding author.}, Dawei Li$^2$, Yalin Wang$^1$\\
$^1$Arizona State University, Tempe, AZ, USA \\
$^2$Samsung Research America, Mountain View, CA, USA\\
{ \{jiezhang.joena, ylwang\}@asu.edu, \{xiaolong.w, dawei.l\}@samsung.com}
}

\begin{document}

\maketitle

\begin{abstract}
Recurrent neural networks (RNNs) achieve cutting-edge performance on a variety of problems. However, due to their high computational and memory demands, deploying RNNs on resource constrained mobile devices is a challenging task. To guarantee minimum accuracy loss with higher compression rate and driven by the mobile resource requirement, we introduce a novel model compression approach \pname{} based on an optimized fast dictionary learning algorithm, which 1) dynamically mines the dictionary atoms of the projection dictionary matrix within layer to adjust the compression rate 2) adaptively changes the sparsity of sparse codes cross the hierarchical layers. Experimental results on language model and an ASR model trained with a 1000h speech dataset demonstrate that our method significantly outperforms prior approaches. Evaluated on off-the-shelf mobile devices, we are able to reduce the size of original model by eight times with real-time model inference and negligible accuracy loss.
\end{abstract}
\section{Introduction}
Deep learning technique is becoming the dominant force of recent breakthroughs in artificial intelligence area~\cite{geng2017novel}. Recently, Recurrent Neural Networks (RNNs) have gained wide attentions with dramatic performance improvements in sequential data modeling, e.g., automatic speech recognition (ASR)~\cite{sak2014long}, language modeling~\cite{lm2}, image captioning~\cite{vinyals2017show}, neural machine translation~\cite{nml1}, etc. Recently, due to the success of RNN-ASR~\cite{lei2013accurate}, personal assistant system~(e.g., Amazon's Alexa, Apple's Siri, Google Now, Samsung's Bixby, Microsoft's Cortana) has become a standard system configuration of smartphones. 

Generally, the trained deep learning models are deployed on the cloud which requires strict Internet connection and also may compromise user privacy. Therefore, there is a high demand to deploy such RNNs with millions of parameters on mobile devices. Deep model compression techniques have been proposed to solve the mobile deployment problem. Specifically for RNNs model compression, a pruning based method ~\cite{narang2017exploring} was introduced which progressively prunes away small parameters using a monotonically increasing threshold during training. Another representative method is to apply singular value decomposition (SVD) low rank approach for decomposing RNNs weight matrices~\cite{prabhavalkar2016compression}. Although some good results for compressing RNNs have been reported, these works fail to consider the hierarchical changes of weight matrices. In addition, none of the existing methods can dynamically adjust the compression rate according to the requirement of the deployed mobile devices. 

\begin{figure}[t]
\centering
\includegraphics[height=5cm]{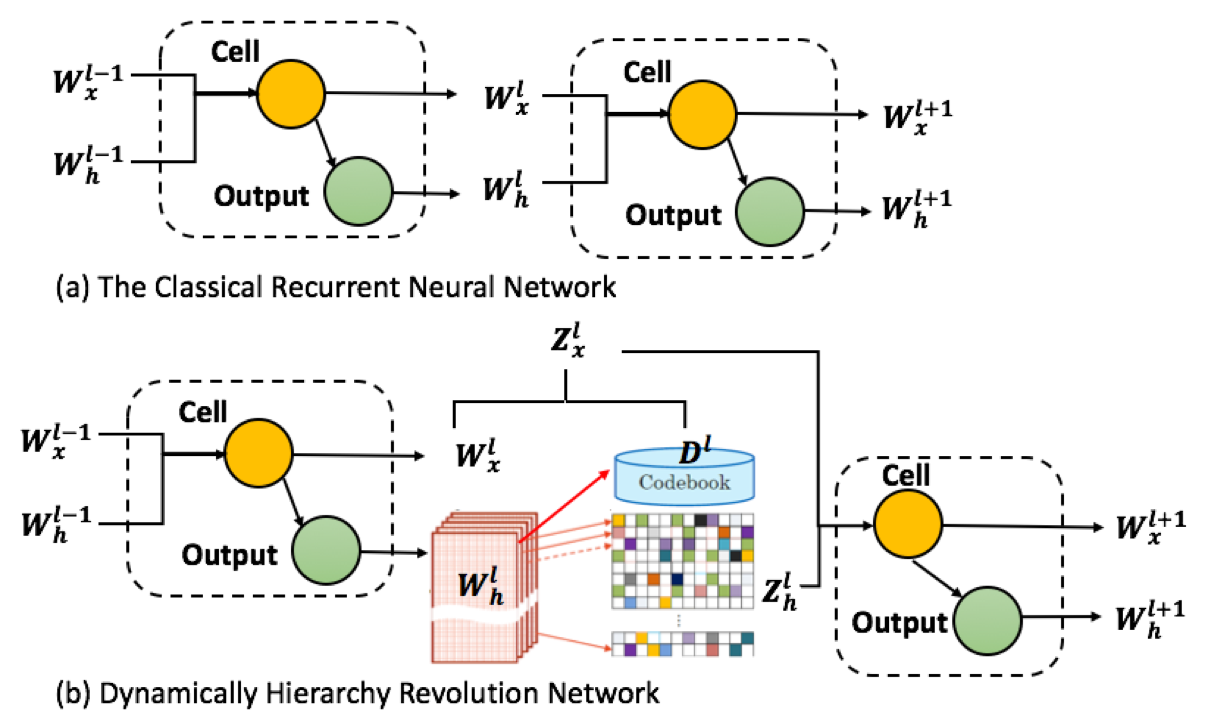}
\caption{(a) The initial model is compressed by (b) jointly dynamically adjust sparsity of recurrent and inter-layer matrices, using a shared projection dictionary.}
\label{fig:1}
\end{figure}

In this work, we propose a dynamically hierarchy revolution (\pname{}) to address the problems in existing RNNs model compression methods (Fig.~\ref{fig:1}). To further compress the model with considering different degrees of redundancies among layers, we exploit a novel way of dynamically mining dictionary atoms from original network structures without manual setting the compression rate for different layers. Our approach achieves significant improvement in model compression compared to previous approaches. It also has much higher re-training's convergence rate and lower accuracy loss compared to the state-of-the-art. In addition, considering different functionalities among hierarchical hidden layers, we come up with the idea of adaptively adjusting sparsity of different hidden layers in RNNs. This can improve the precise structure approximation to the original network with minimizing the performance drop.

Our most significant contributions can be summarized into threefold: Firstly, this is the first work dynamically exploring dictionary learning in weight matrix of both RNN and LSTM that jointly compresses both hidden layer and inter layer to reconstruct original weight matrices. Secondly, our approach is a dynamic model compression algorithm. It can dynamically mine the dictionary atoms of the projection dictionary matrix within the layer to better learn a common codebook representation across inter-layer and recurrent layer.
This can find the optimal numbers of neurons among layers to better control the degree of compression with negligible accuracy loss.
Thirdly, given a hierarchical RNN structure, our approach can adaptively set various sparsities of sparse codes for each layer to minimize the performance drop for the compressed network. The experimental results demonstrate that \pname{} achieves significant improvement in terms of  compression scale and speedup rate compared to the state-of-the-art approaches.

\section{Related Work}
In general, we can summarize existing RNN compression methods into three categories: Pruning, low-rank approximation, and knowledge distillation. The first pruning work was proposed by Han et al.~\shortcite{han2015learning} where a hard threshold was applied as the pruning criterion to compress the deep neural network. Li et al.~\shortcite{li2017deeprebirth} discussed the optimization of non-tensor layers such as pooling
and normalization without tensor-like trainable parameters for deep model compression.
Xue et al.~\shortcite{xue2013restructuring} used low-rank approximation to reduce parameter dimensions in order to save storage and reduce time complexity. Denile et al.~\shortcite{denil2013predicting} showed that the given neural network can be represented by a small number of parameters. Sainath et al.~\shortcite{sainath2013low} reduced the number of model parameters by applying the low-rank matrix factorization to simplify the neural network architecture. 

Hinton et al~\shortcite{hinton2015distilling} learned a small student network to mimic the original teacher network by minimizing the loss between the student and teacher output. Kim et al.~\shortcite{kim2016sequence} integrated knowledge distillation to approximately match the sequence-level distribution of the teacher network. Chen et al.~\shortcite{chen2017darkrank} introduced a knowledge of cross sample similarities for model
compression and acceleration.

In addition, there are some other compression methods. Low-precision quantization~\cite{xu2018alternating} is a scalar-level compression method without considering the relation among learned parameters in layers. 
Compared to these works, our approach dynamically adjusts the compression rate across layers and explores the sparsity of weight matrices to compress RNN. It achieves better compression rate with less accuracy loss compared with above methods.

Recently, Han et al.~\shortcite{han2017ese} used a pruning method to compress the LSTM on speech recognition problem and it reduced the parameters of the weight matrix to 10\% and achieved 5X compression rate (each non-zero element is represented by its value and index in the matrix). However, DirNet considers the sparsity of various matrices and dynamically adjusts the sparsity based on the feature of models and gets a better compression rate. Our approach can compress the model by around 8X with negligible performance loss. 
\section{Methods}
In this section, we first introduce the background of RNNs. Then, we present the basic sparse coding based RNN compression method without considering the mobile requirement and the hierarchical network structure. Further, we present the proposed approach \pname{}.  
 
\subsection{RNN Background}
Let $t = 0, 1, \cdots, T$ denotes time steps and $l = 1, 2, \cdots, L$ denotes the hidden layers in RNN. At time $t$, $h_t^l \in \mathbb{R}^{N^l}$ denotes the activations of the $l$-th hidden layer with $N^l$ node. Therefore, the inputs to this layer at time $t$ are donated by $h_t^{l-1} \in \mathbb{R}^{N^{l-1}}$. Then, we can define the output activations of the $l$-th and $(l+1)$-th layers in a classical RNN:
\begin{equation}
\label{eq:2}
h_t^l = \sigma(W_x^{l-1}h_t^{l-1} + W_h^lh_{t-1}^l + b^l),
\end{equation}
\begin{equation}
\label{eq:3}
h_t^{l+1} = \sigma(W_x^{l}h_t^{l} + W_h^{l+1}h_{t-1}^{l+1} + b^{l+1}),
\end{equation}
where $\sigma(\cdot)$ denotes a non-linear activation function and $b^l \in \mathbb{R}^{N^l}$ and $b^{l+1} \in \mathbb{R}^{N^{l+1}}$ represent bias vectors. $W_x^l \in \mathbb{R}^{N^{l+1}\times{N^l}}$ and $W_h^l \in \mathbb{R}^{N^{l}\times{N^l}}$ denote inter-layer and recurrent weight matrices, respectively. 

\begin{figure}[t]
\centering
\includegraphics[height=5cm]{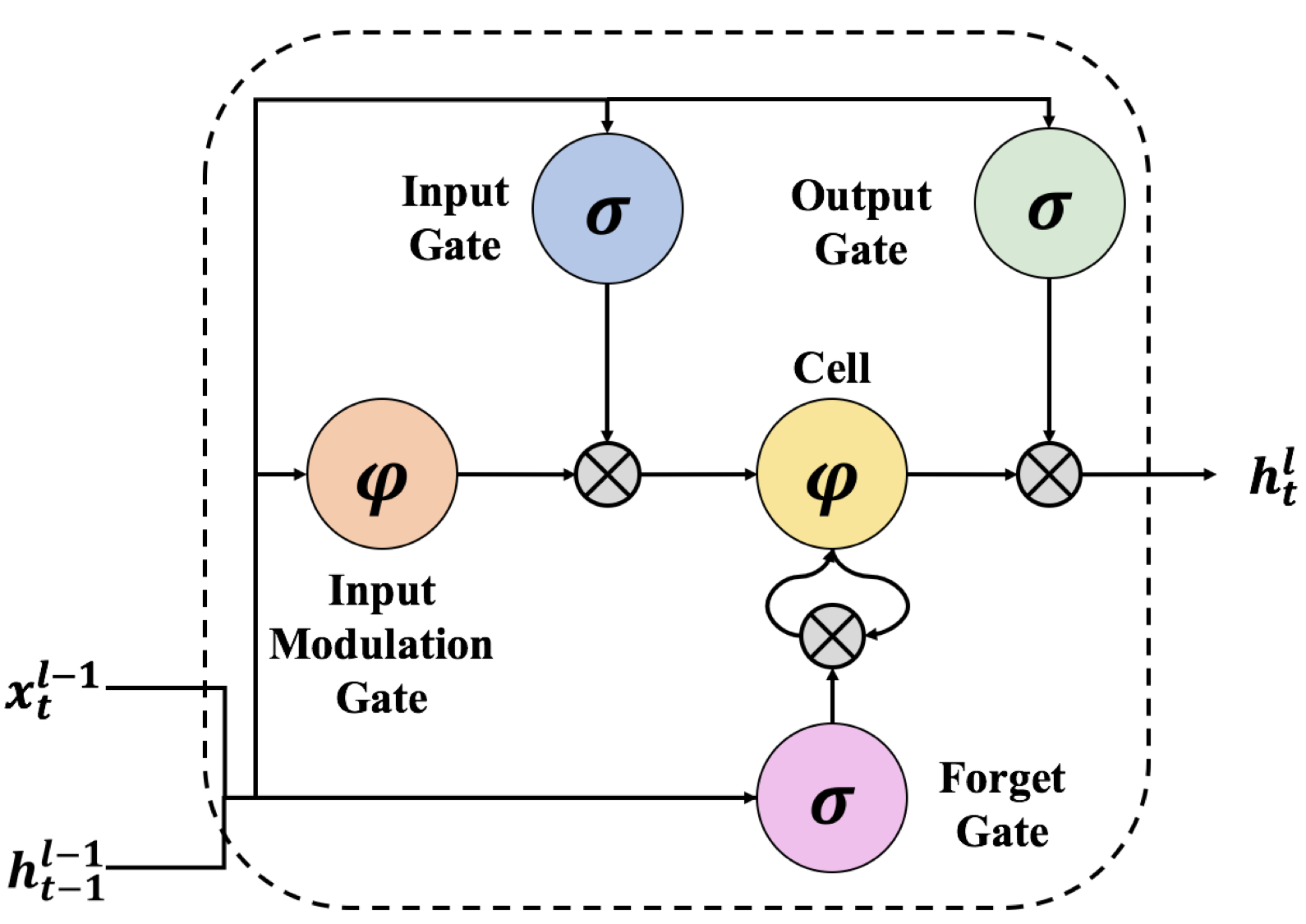}
\caption{A graphical representation of LSTM memory cells used in this work.}
\label{fig:2}
\end{figure}

Different from traditional RNNs, LSTM contains specialized cells called \emph{memory cell} in the recurrent hidden layer. Empirically, LSTM has a memory cell for storing information for long periods of time, we use $c_t^l\in\mathbb{R}$ to denote the long-term memory. LSTM can decide to overwrite the memory cell, retrieve it, or keep it for the next time step~\cite{zaremba2014recurrent}. In this paper, our LSTM architecture is the same as~\cite{graves2013speech}. The general structure is illustrated in Figure~\ref{fig:2}.

\begin{equation}
\label{eq:4}
\left(\begin{array}{c} i \\ o \\ f \\ g \end{array} \right) = \left( \begin{array}{c} sigmod \\ sigmod \\ sigmod \\ \tanh \end{array} \right)  (T_{n, 4n}, T_{n, 4n}) \left( \begin{array}{c} h_t^{l-1} \\ h_{t-1}^l \end{array} \right) ,
\end{equation}
\begin{equation*}
c_t^l = f \odot c_{t-1}^l + i \odot g, h_t^l = o \odot \tanh(c_t^l),
\end{equation*}
where $i, o, f$ and $g$ are input gate, output gate, forget gate and input modulation gate, respectively. And $c$ denotes the memory cells and all gates are the same size as the memory cells. $\odot$ is the element-wise product of the vectors. $n$ and $4n$ denote the dimension of one gate and four gate, respectively. 

\subsection{Sparse Coding based Model Compression}
As indicated above, one direct way to compress RNN is to sparsify weight matrix $W_x^l$ and $W_h^l$. Given a weight matrix $W \in \mathbb{R}^{p\times t} $,  each $w_i \in \mathbb{R}^p$. Sparse coding aims to learn a dictionary $D$ where $D \in \mathbb{R}^{p\times m}$ and a sparse code matrix $Z \in \mathbb{R}^{m\times t}$. The original matrix $W$ is modeled by a sparse linear combination of $D$ and $Z$ as $W\approx DZ$. We can formulate the following optimization problem:
\begin{equation}
\min_{D\in \Psi, z_i \in \mathbb{R}^{l}} f(D, Z)  = \frac{1}{2}||W-DZ||^2_F+\lambda||Z||_1,
\label{eq:1}
\end{equation}
where $\Psi = \{ D \in \mathbb{R}^{p\times m}: \forall j\in 1, \cdots, m, ||d_j||_2 \leq 1 \}$, $d_j$ denotes the $j$th column of $D$ and $Z=(z_1,\cdots, z_t)$. $\lambda$ is the positive regularization parameter and $||W||_F$ denotes the Frobenius norm of the matrix. 

For RNNs, we can compress simultaneously the weight matrices of both recurrent layer and inter-layer (from Eq.~\eqref{eq:2} or Eq.~\eqref{eq:3}). 
Such joint compression can be achieved by a projection matrix~\cite{sak2014long}, denoted by $D^l \in \mathbb{R}^{p^l\times N^l}$ ($p^l < N^l$), such that, $W_h^l = D^lZ_h^l$ and $W_x^l=D^lZ_x^l$. Therefore, we can incorporate this idea into Eq.~\eqref{eq:2} and Eq.~\eqref{eq:3}, then we get the followings:
\begin{equation}
\label{eq:9}
\min_{D^l, Z_h^l} ||W_h^l-D^lZ_h^{l}||+\lambda_1|Z_h^l|_1,
\end{equation}
\begin{equation}
\label{eq:99}
h_t^l = \sigma(W_x^{l-1}h_t^{l-1} + D^lZ_h^lh_{t-1}^l + b^l),
\end{equation}
\begin{equation}
\label{eq:10}
\min_{Z_x^l} ||D^lZ_x^l - W_x^l||_2^2+\lambda_2||Z_x^l||_1,
\end{equation}
\begin{equation}
\label{eq:100}
h_t^{l+1} = \sigma(D^lZ_x^lh_t^{l} + W_h^{l+1}h_{t-1}^{l+1} + b^{l+1}),
\end{equation} 
where $Z_h^l\in\mathbb{R}^{p^l\times N^l}$ and $Z_x^l\in\mathbb{R}^{p^l\times N^{l+1}}$. Therefore, we can calculate the compression rate by $\frac{N^l \times N^l + N^{l+1} \times N^l}{p^l\times N^l + p^l\times N^l + p^l\times N^{l+1}}.$

However, a problem exists in this deep compression model is that the compression rate and the sparsity are manually set which limits the compression space.  It would be much more efficient and promising to automatically set the compression rate and sparsity based on the understanding of deep neural network's structure. Also we should consider minimizing the training time during the model compression which may take weeks or even months, especially for RNN based sequential models. In this paper, we propose \pname{}, which can dynamically adjust the sparsity of Eq.~\eqref{eq:9} and Eq.~\eqref{eq:10}, to meet this requirement. 

\begin{algorithm}[t]
\caption{\pname{}}
\label{alg:1}
\SetKwData{Left}{left}\SetKwData{This}{this}\SetKwData{Up}{up}
\SetKwFunction{Union}{Union}\SetKwFunction{FindCompress}{FindCompress}
\SetKwInOut{Input}{input}\SetKwInOut{Output}{output}
\Input{Inter-layer matrix $W_h$ and recurrent matrix $W_x$, shift operators $\triangle$, $K, M, \lambda_1=0.1, \lambda_2$}
\Output{Compressed weight matrices $Z_h, Z_x$ and projection matrix $D$}
\Begin{
	\For{$l = 1\rightarrow L$}{
  		Initialize $D^l$ and $Z_h^l$\\
       \Repeat{convergence}{  		
  		Update sparse code $z_i^k\leftarrow CD(d_i^k, z_i^{k-1}, w_i)$\\
  		Update dictionary $d_k\leftarrow \arg\min_{\{d_i\}_{i=1}^K}\sum_{j=1}^M \frac{1}{2}||x_j - \sum_{i=1}^Kz_{ij}\delta_{ij}(d_i)||_2^2, s.t. ||d_k||_2=1$\\
       }  		
       $h_t^l = \sigma(W_x^{l-1}h_t^{l-1} + D^lZ_h^lh_{t-1}^l + b^l)$\\
  		\While{model is unempty}{
     		$t=0$\\
    		\Repeat{Convergence}{
               $t = t + 1$ \\        			
       			Fix $\Theta_t^l$ update $z_t^l$ according to Eq.~\ref{eq:12}\\
       			Fix $z_t^l$ update $\Theta_t^l$ according to Eq.~\ref{eq:5}\\
    		}
    		$\Theta_0^{\tau+1} = \Theta_t^{\tau}$\\
    		$\tau = \tau + 1$
  	  }
  	  $h_t^{l+1} = \sigma(D^lZ_x^lh_t^{l} + W_h^{l+1}h_{t-1}^{l+1} + b^{l+1})$
  }
   \KwRet{$Z_h, Z_x, D$}\;
}
\end{algorithm}

\pname{} is mainly composed of two steps and depicted graphically in Fig.~\ref{fig:1}: In Step 1, a dynamically adaptive dictionary learning (Sec.~\ref{dynamic}) is proposed to adapt the atoms in shared dictionary among inter-layer and recurrent layer to adjust the sparsity (compression rate and accuracy) within the layer (Eq.~\eqref{eq:9}). In Step 2, we propose an adaptive sparsity learning approach with considering the network's hierarchical structure (Sec.~\ref{adaptive}) to optimize the $Z$ and sparsity parameter $\Theta$ hierarchically with an appropriate sparsity degree. The sparsity (compression rate and accuracy) achieved by different layers is depending on the number of neurons and the architecture of the specific layer. In general, the non-zero values in $Z$ are ten to twenty percent of the original weight matrices $W$ in our work. From this, we can find the proposed approach adaptively set various sparsity among the hierarchical RNN architecture on Eq.~\ref{eq:10} instead of using a fixed fraction function of explained variance as advocated in Prabhavalkar et al.~\shortcite{prabhavalkar2016compression}. We summarize the key steps of \pname{} in Algorithm~\ref{alg:1}.

\subsection{Dynamically Adaptive Dictionary Learning}
\label{dynamic}
It is an essential step to dynamically adjust the dimension of the projection matrices $D^l$ to get the optimal compression rate. We introduce a shift operation $\triangle$ on atoms to better adjust the compression rate. Given a set of shift operations $\triangle$ which contains only small shifts relative to the size of the (time window), for every $j$ there exist coefficients $z_{ij} \in\mathbb{R}$ and shift operators $\delta_{ij}\in\triangle$~\cite{hitziger2013jitter}, such that $w_j = \sum_{i=1}^K z_{ij}\delta_{ij}(d_i).$

Now, we can formulate the dynamically adaptive dictionary learning problem as follows:
\begin{equation}
\min_{d_i, z_{ij}, \delta_{ij}} \sum_{j=1}^M(\frac{1}{2}||W_h^l-\sum_{i=1}^Kz_{ij}\delta_{ij}(d_i)||_2^2+\lambda_1||z_j||_1),
\end{equation}
\begin{equation*}
s.t.\quad ||d_i||_2 = 1, \delta_{ij}\in\triangle, K = N^l, M = p^l.
\end{equation*}
The problem becomes Eq.~\ref{eq:1} when $\triangle = \{I\}$, thus we use alternate minimization to solve it.


\textbf{Sparse Codes Update}
It is known that updating the sparse code is the most time consuming part~\cite{mairal2009online}. One of the state-of-the-art methods for solving such lasso problem is Coordinate descent (CD)~\cite{friedman2007pathwise}. Given an input vector $w_i$, CD initializes $z_i^0 = 0$ and then updates the sparse codes many times via matrix-vector multiplication and thresholding. However, the iteration takes thousands of steps to converge. Lin et al.~\shortcite{lin2014stochastic} observed that the support (non-zero element) of the coordinates after less than ten steps of CD is very accurate. Besides, the support of the sparse code is usually more important than the exact value of the sparse code. 
Therefore, we update the sparse code $z_i$ by using a few steps of CD operation because the original sparse coding is a non-convex problem and do not need to run CD to the final convergence. For the $k$-th epoch, we denote the updated sparse code as $z^k_i$. It will be used as an initial sparse code for the $k+1$-th epoch.

Update $z_i^k$ via one or a few steps of coordinate descent:
\begin{equation*}
z_i^k \leftarrow  CD(D^k_i, z_i^{k-1}, x_i).
\end{equation*}
Specifically, for $j$ from 1 to $m$, we update the $j$th coordinate $z_{i, j}^{k-1}$ of $z_i^{k-1}$ cyclicly as follows:
\begin{equation*}
b_j \leftarrow (d_{i, j}^k)^T(w_i - d_i^kz_i^{k-1}) + z_{i, j}^{k-1},  
\end{equation*}
\begin{equation*}
z_{i, j}^{k-1} \leftarrow s_{\lambda}(b_j),
\end{equation*}
where $s$ is the soft thresholding shrinkage function~\cite{combettes2005signal}. We call above updating cycle as one step of CD. The updated sparse code is then denoted by $z_i^k$. 

\textbf{Dictionary Update}
For updating the dictionary, we use block coordinate descent~\cite{tseng2001convergence} for updating each atom $d_k$. 
\begin{equation*}
 d_k = \arg\min_{d_k} \sum_{j=1}^M\frac{1}{2}||w_{j}^l - \sum_{i=1}^Kz_{ij}\delta_{ij}(d_i)||_2^2, s.t. ||d_k||_2 = 1.
\end{equation*}

This can be solved in two steps, the solution of the unconstrained problem by differentiation followed by normalization and can be summarized by 
\begin{equation}
\label{eq:7}
\tilde{d_k} = \sum_{j=1}^M z_{kj}\delta^{-1}_{kj}(w_{j}^l - \sum_{i\neq k} z_{ij}\delta_{ij}(d_i)),
\end{equation}
\begin{equation*}
d_k = \frac{\tilde{d_k}}{||\tilde{d_k}||_2}.
\end{equation*}

As in~\cite{mairal2009online}, we found that one update loop through all of the atoms was enough to ensure fast convergence of Algorithm~\ref{alg:1}.  The only difference of this update compared to common dictionary learning is the shift operator $\delta_{ij}$. In Eq.~\eqref{eq:7}, $\delta\delta^t = I$. If the shift operator is a non-circular operators,  the inverse $\delta^{-1}_{kj}$ needs to be replaced by the adjoint $\delta^t_{kj}$ and the rescaling function $\phi = (\sum_{j=1}^M z_{kj}^2\delta_{kj}\delta^t_{kj})^{-1}$ needs to be applied to the update term. 

Besides, we used a random selection method (randomly select $\hat{l}$ samples from matrices $W_h^l$ to construct initial dictionaries $D^l$) to initialize the dictionaries for different layers in \pname{}. Then, we set all the sparse codes $Z_h^l$ to be zero in the beginning and $k = 10$ epoch. 

\subsection{Adaptively Hierarchical Network}
\label{adaptive}
Considering different functionalities among hierarchical hidden layers, we propose the method of adaptively changing sparsity in different hidden layers. We use an initial perturbations as ~\cite{zhang2016annealed} to let all features can be selected by competing with each other. Then we gradually shrank the network by using stronger $l_1$-penalties and fewer features remaining in the progressive shrinking. Therefore, \pname{} will go through the self-adjusting sequential stages before reaching the final optimal. 

We note that sharing $D^l$ across the inter-layer and recurrent-layer matrices allows more efficient parameterization for the weight matrices. Besides, this does not result in a significant loss of performance. By adjusting the dimensions of the projection matrices ($p^l$) in each of the layers of the network, the compression rate of the model will be determined. Therefore, after fixing the dictionary to $D^l$, solving Eq.~\ref{eq:10} is equivalent to solve a LASSO problem~\cite{tibshirani1996regression}. The objective function of solving $Z_x^l$ as follow:
\begin{equation}
\label{lasso}
\min_{Z_x^{l}} ||W_h^{l+1} - D^lZ_x^{l}||_2^2 + \lambda_2|Z_x^{l}|_1.
\end{equation} 

After we get dictionary $D^l$ from Eq.~\ref{eq:9}, we can determine $Z_x^l$ as the solution to the following LASSO problem with adaptive weights to regularize the model coefficients along with different features:
\begin{equation}
\min_{Z_h^{l+1}} ||W_h^{l+1} - D^lZ_h^{l+1}||_2^2 + \lambda_2|\Theta\odot Z_h^{l+1}|_1,
\end{equation}
where $\lambda_2$ is different across layers and we selected the best value from $10^-3$ to $10^3$ in this work. $\Theta$ denotes regularization weight, we will receive different penalized matrices $Z_h^{l+1}$ instead of controlling the sparsity by $\lambda_2$ as in Eq.~\eqref{lasso}. In addition, $|\Theta\odot Z_h^{l+1}|_1=\sum_i\theta_i |z_i^{l+1}|$ and $\Theta = |Z_{ols}|^{-\gamma}$, where $Z_{ols}$ is the ordinal least-square solution. 

Therefore, we can consider the following linear regression problem with the inter weight matrix $W_x^l \in \mathbb{R}^{ N^{l+1} \times N^l}$, where $N^{l+1}$ is the sample size and $N^l$ is the dimension of the target response vector. We use $n$ to represent $N^{l+1}$. Then, we use an adaptive weight vector $\Theta = [\theta_1, \theta_2, \cdots, \theta_n]^T \in \mathbb{R}^n$ to regularize over different covariates, as
\begin{equation}
\label{eq:11}
\min_{\Theta, Z_x} ||W_x^l - D^lZ_x^l||_2^2 + \lambda_2||\Theta^{-\gamma}\odot Z_x^l||_1,
\end{equation}
\begin{equation*}
s.t. \sum_i \theta_i = \vartheta, \theta_i \geq 0,
\end{equation*}
where $||\Theta^{-\gamma} \odot Z_x^l||_1 = \sum_{i=1}^n \theta_i^{-\gamma}\cdot |Z_x^l|$ and we alternatively optimize $\Theta$ and $Z_x^l$ in the learning process. 

Suppose we initialize $\Theta^l$, with $|\Theta^l| = \Theta^l_0$, $l$ denotes $l$-th layer. Then, we alternatively update $\Theta^l$ and $Z_x^l$ in Eq.~\eqref{eq:11} under this equality norm constraint until convergence. We will start the second stage of iterations with an updated norm constraint $|\Theta^l| = \Theta^l_1$ after the initial stage, which imposes a stronger $l_1$ penalty. Then we alternatively update $\Theta^l$ and $Z_x^l$ until the second stage ends. We keep strengthening the global $l_1$-norm regularization stage by stage during the updating procedure. We use $\tau$ to denote the index of each stage and $\Theta^l = \Theta_{\tau}^l$ is the compose of iteration.

To solve Eq.~\eqref{eq:11}, we first fix $\Theta$ and solve $Z_x^l$, which can be computationally converted to a LASSO problem: 
\begin{equation}
\label{eq:12}
\min_{Z_x^l} ||W_x^l - D^lZ_x^l||_2^2 + \lambda_2||\Theta^{-\gamma}\odot Z_x^l||_1.
\end{equation}

Then, when we fix $Z_x^l$ update $\Theta$, the problem becomes the following constrained optimization problem:
\begin{equation}
\label{eq:5}
\min_{\Theta} \sum_i z_i \cdot \theta_i^{-\gamma},
\end{equation}
\begin{equation*}
s.t. \sum_i \theta_i = \vartheta, \theta_i \geq 0.
\end{equation*}
We used the Lagrangian of Eq.~\ref{eq:5}, and drop the non-negativity constraint. Let $c_i = |z_i|_1$. Then the Lagrangian can be written as
\begin{equation*}
J = \sum_i\theta_ic_i^{-\gamma} + \beta(\sum_ic_i-\vartheta). 
\end{equation*} 
By setting $\partial{J}/\partial{\theta_i} = 0$, we have
\begin{equation}
\label{eq:13}
\beta = \frac{c_{i\gamma}}{\theta_i^{1+\gamma}}.
\end{equation}
Plugging the above relation in the constraint $\sum_i\theta_i = \vartheta$, then we have 
\begin{equation*}
\theta_i^{1+\gamma} = \frac{c_{i\gamma}}{\beta} = \frac{c_{i\gamma}\cdot \vartheta^{1+\gamma}}{(\sum_i(c_{i\gamma})^{\frac{1}{1+\gamma}})^{1+\gamma}}.
\end{equation*}
Finally, we plug the above equation into Eq.~\ref{eq:13} and get
\begin{equation}
\label{eq:6}
\theta_i = (\frac{c_i^{\frac{1}{1 +\gamma}}}{\sum_{i=1}^nc_i^{\frac{1}{1 +\gamma}}})\vartheta.
\end{equation}
Since $c_i = \sum_j|z_j^i| \geq 0$ and $\vartheta \geq 0$, the solution will satisfy the non-negative constraints automatically. After we trained the original RNN model and retrieved the weight matrices of both inter-layer and recurrent layer, the proposed algorithm is applied to learn the new matrices $Z_h, Z_x$ and $D$ of each layer, which are used to initialize parameters of neural network layers in the new network structure (Fig.1 (b)). After the initialization, we fine-tune the neural network as Han et al.~\shortcite{han2015deep} and the sparsity is preserved by only updating non-zero elements in the sparse matrix. Fine-tune step is also required for other compression works, e.g., \cite{prabhavalkar2016compression}.

\subsection{Extended Model Compression}
We extend the Sec.~\ref{dynamic} - Sec.~\ref{adaptive} from standard RNNs to LSTM. In LSTM, the recurrent-weight matrix $W_h^l$ and the inter-layer matrix $W_x^l$ are both concatenation of four gates, which are input gate, output gate, forget gate and input modulation gate. We stack them vertically and denote as $(i, o, f, g)^T$.

In this work, we do not consider the peephole weights because it already narrows and will not compress a lot of parameters. Thus, we rewrite the Eq.~\eqref{eq:2} and Eq.~\eqref{eq:3} as follows:
\begin{equation*}
\left(\begin{array}{c} i \\ o \\ f \\ g \end{array} \right) = \left( \begin{array}{c} sigmod \\ sigmod \\ sigmod \\ \tanh \end{array} \right)  (T_{n, 4n}, D^l_{n, 4n}Z^l_{h(n, 4n)}) \left( \begin{array}{c} h_t^{l-1} \\ h_{t-1}^l \end{array} \right) 
\end{equation*}
\begin{equation*}
\left(\begin{array}{c} i \\ o \\ f \\ g \end{array} \right) = \left( \begin{array}{c} sigmod \\ sigmod \\ sigmod \\ \tanh \end{array} \right)  (D^l_{n, 4n}Z^{l+1}_{x(n, 4n)}, T_{n, 4n}) \left( \begin{array}{c} h_t^{l} \\ h_{t-1}^{l+1}\end{array} \right) 
\end{equation*}

Besides, compressing a single-layer LSTM model is a special case for the proposed DirNet: (1) we can still learn the shared $D^l$ using $W_x^l$ and $W_h^l$ of the single LSTM/RNN layer as in Fig.1 (b) to achieve a high compression rate on a single-layer LSTM model; (2) compressing multiple LSTM/RNN layers, which can adaptively change the sparsity of the sparse codes relying on cross-layer information, will achieve higher compression rate than a single-layer LSTM/RNN model.

\section{Experimental Results and Discussion}
The proposed approach is a general algorithm for compressing recurrent neural networks including vanilla RNN, LSTM and GRU, etc, and thus can be directly used in many problems, e.g., language modeling (LM), speech recognition, machine translation and image captioning. In this paper, we select two popular domains: LM and speech recognition. We compare \pname{} with (1) \textit{LSTM-SVD} and \textit{LSTM-ODL}. \textit{LSTM-SVD} is the method proposed by Prabhavakar et al.~\shortcite{prabhavalkar2016compression} which compresses RNNs by low-rank SVD. \textit{LSTM-ODL} is the method which compresses RNNs by online dictionary learning~\cite{mairal2009online}. To train 
\pname{}, it takes 10 hours on PTB with 1 K80 GPU and around 120 hours on LibriSpeech using 16 Nvidia K80 GPUs. 

\subsection{Language Modeling}
We conduct word-level prediction on the Penn Tree Bank (PTB) dataset~\cite{marcus1993building}, which consists of 929k training words, 73k validation words and 82k test words. We first train a two-layer unrolled LSTM model with 650 units per layer, which is using the same network setting of LSTM-medium in~\cite{zaremba2014recurrent}. For comparison, the same model is compressed using LSTM-SVD, LSTM-ODL as well as \pname{}. We report the result of testing data in Table~\ref{table:1}. We observe that \pname{} achieves superior performances regarding compression rates and speedup rates while maintaining the accuracy. The reason is that dynamically adjusting the projection matrices $D$ can achieve better compression rate and adapt different sparsities across layers can receive negligible accuracy loss.

\begin{table}[t]
\centering
 \begin{tabular}{c|c|c|c|c}
\hline
Network & $\#$ Params & R & T & PER  \\ \hline
LSTM &4.7M&1x&1x&81.3 \\\hline
LSTM-SVD &2.4M&2.0x&1.9x&81.4\\
LSTM-ODL &1.6M&2.9x&2.8x&81.5\\
\pname{} &0.7M&6.7x&6.4x&81.5\\\hline
LSTM-SVD &0.7M&6.7x&6.4x&87.4\\
LSTM-ODL &0.7M&6.7x&6.4x&83.2\\
\pname{} &0.7M &6.7x&6.4x&81.5\\\hline
\end{tabular}
\caption{Comparison of the number of parameters (Params), compression rates (R), speedup on mobile CPU speed (T) and Perplexity (PER) on PTB dataset. }
\label{table:1}
\end{table}

\subsection{Speech Recognition}
\textbf{Dataset:} The LibriSpeech corpus is a large (1000 hour) corpus of English read speech derived from audiobooks in the LibriVox project, sampled at 16kHz.  The accents are various and not marked, but the majority are US English. It is publicly available at http://www.openslr.org/12/~\cite{panayotov2015librispeech}. LibriSpeech comes with its own train, validation and test sets and we use all the available data for training and validating our models. Moreover, we use the 100-hour ``test clean" set as the testing set. Mel-frequency cepstrum (MFCC)~\cite{davis1990comparison} features are computed with 26 coefficients, a 25 ms sliding window and 10 ms stride. We use nine time slices before and nine after, for a total of 19 time points per window. As a result, with 26 cepstral coefficients, there are 494 data points per 25 ms observation.

\textbf{Baseline Model:} Following~\cite{prabhavalkar2016compression}, our baseline model for speech recognition is a 5-layer RNN model with the Connectionist Temporal Classification (CTC) loss function~\cite{graves2006connectionist}, which predicts 41 context-independent (CI) phonemes~\cite{sak2015learning}. Each hidden RNN layer is composed of 500 LSTM units. We train all models for 100 epochs using Adam optimizer~\cite{adamopt} with an initial learning rate of 0.001. It takes around 35 and 55 epochs to converge of \pname{} and the baseline model, respectively. The size of the batch is 32. The model is first trained to convergence for optimizing the CTC criterion, followed which are sequences
discriminatively trained to optimize the state-level minimum Bayes risk (sMBR) criterion~\cite{kingsbury2009lattice}. We implement \pname{} by Tensorflow~\cite{tensorflow} and use Samsung Galaxy S8 smartphone as the mobile platform to evaluate the performance. 

\begin{table}[t]
\centering
 \begin{tabular}{c|c|c|c|c}
\hline
Network & $\#$ Params & R & T & WER  \\ \hline
LSTM &10.3M&1x&1x&12.7 \\\hline
LSTM-SVD &6.9M&1.5x&1.4x&12.7\\
LSTM-ODL &3.7M&2.8x&2.6x&12.7\\
\pname{} &2.4M&4.3x&4.2x&12.7\\\hline
LSTM-SVD &1.3M&7.9x&7.6x&16.1\\
LSTM-ODL &1.3M&7.9x&7.6x&14.3\\
\pname{} &1.3M &7.9x&7.6x&12.9\\\hline
\end{tabular}
\caption{Comparison of the number of parameters (Params), compression rates (R), speedup on mobile CPU speed (T) and Word error rates (\%) (WER) on LibriSpeech dataset. }
\label{table:2}
\end{table}

\textbf{Results:}
To give a comprehensive evaluation of the proposed approach, we list the word error rate along with parameter number and the execution time comparison between different approaches in Table~\ref{table:2}. There are two sets of experiments: 1) compare the maximum compression rate obtained by different approaches without the accuracy drop 2) measure the word error rate among different approaches with the same compression rate.

As indicated in the first row of Table ~\ref{table:2}, the basic LSTM model can achieve 12.7\% (WER) before any compression. It includes 10.3M parameters in total and takes 1.77s to recognize 1s audio. Our proposed approach \pname{} can compress the model by 4.3 times without losing any accuracy. The speed up achieved by \pname{} is 4.2 times faster compared to 1.4 and 2.6 times obtained by LSTM-SVD and LSTM-ODL. These results indicate the superiority of the proposed \pname{} over current compressing algorithms. Moreover, such significant improvement shows that dynamically adjust sparsity across layers might help the compression models receive higher compression rates. From the results listed in the second row of Table~\ref{table:2}, we can find that when compressing the model size by 7.9 times, LSTM-ODL and LSTM-SVD lose lots of accuracy, especially SVD based compression algorithm, which is 3.4\% (16.1\%-12.7\%) compared to 0.2\% achieved by \pname{}.

\subsection{Adaptively Shrinking Parameter Selection}
In this section, we study how the performance of our approach is affected by the following two parameters: $\Theta^0$ that controls the initial ``sparsity'' of the system and the shrinking factor $\gamma$ that controls the compression rate of the system. We use F-score on LibriSpeech dataset to measure the performance. 

\begin{figure}[t]
\centering
\includegraphics[height=5cm]{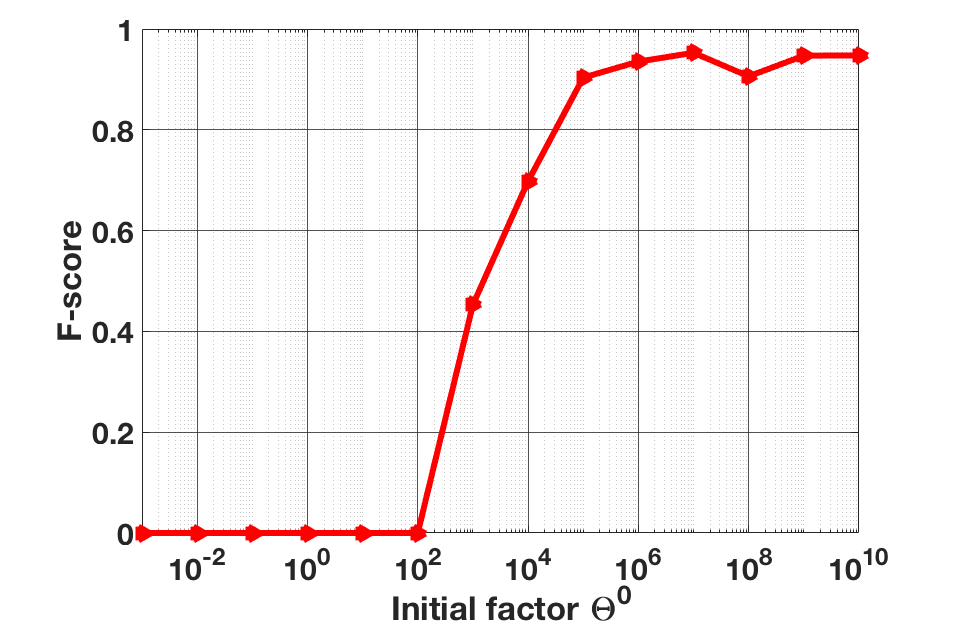}
\includegraphics[height=5cm]{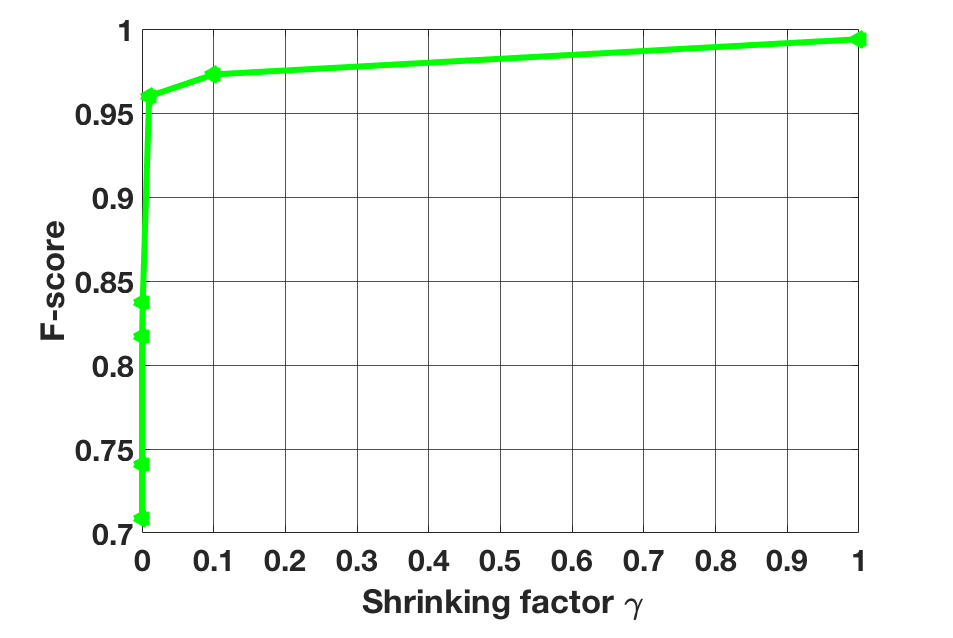}
\caption{Performance of different parameter selections.}
\label{fig:3}
\end{figure}

First, we examine the performance of choosing different $\Theta^0$ values. The range is from $10^{-3}$ to $10^{10}$, and the result is shown on the left side of the Fig.~\ref{fig:3}. We notice that when $\Theta^0$ is below $10^5$, the performance quickly drops and the lower initial sparsity even fails to start the whole system with sufficient energy. As a result the iterations could quickly stop at a local optimal. Therefore, we choose the best performance $\Theta^0 = 10^7$ in all experimental settings.

Second, we study the influence of varying shrinking factor $\gamma$. The explored value range is from $10^{-6}$ to $10^{0}$. We observe that the performance sharply increases first and becomes stable afterward. Besides, the system evolves slowly such that the shrinking stage is sufficient when the shrinking scheme $\gamma \rightarrow 1$. In practice, we choose $\gamma = 0.4$ to strike a balance between efficiency and the quality of shrinking procedure.

\subsection{Adaptive Hierarchy of the Network}
We also compare the performance of \pname{} (O) adaptive cross-layers' compression capability with LSTM-SVD (G).  All the results are listed in Table~\ref{table:3}. We can find that \pname{} achieves much lower WER compared to LSTM-SVD even with the same number of neurons. Also, these results further demonstrate the superiority of \pname{} adaptive adjusting sparse features in deep model compression. 

\begin{table}[t]
\centering
 \begin{tabular}{c|c|c|c}\hline
$\#$ neurons of each layer & Params & WER (G) & WER (O)  \\ \hline
500, 500, 500, 500, 500&10.3M&12.7&12.7 \\ \hline
350, 375, 395, 405, 410&8.6M& 12.3&12.3\\ \hline
270, 305, 335, 345, 350&7.2M&12.5 &12.3\\ \hline
175, 215, 245, 260, 265&5.4M&12.5&12.4\\ \hline
120, 150, 180, 195, 200&4.1M&12.6&12.5 \\ \hline
80, 105, 130, 145, 150&3.1M&12.9&12.5\\ \hline
50, 70, 90, 100, 110&2.3M&13.2&12.7\\ \hline
30, 45, 55, 65, 75&1.7M&14.4&12.9\\ \hline
25, 35, 45, 50, 55&1.3M&16.6&12.9\\ \hline
\end{tabular}
\caption{The Word error rates (\%) (WER) on the testing set as varying the dimensions of each projection matrices $D^l$ by adaptively Adjusting the hierarchical structure of RNN.}
\label{table:3}
\end{table}
\section{Conclusions}
In this paper, we introduce \pname{} that dynamically adjusts the compression rate of each layer in the network, and adaptively change the hierarchical structures among different layers on their weight matrices. Experimental results show that compared to other RNN compression methods, \pname{} significantly improves the performance before retraining. In our ongoing work, we will integrate scalar level compression with our \pname{} to further compress the deep model.

\vspace{-0.5em}
\bibliographystyle{named}
\bibliography{egbib}
\end{document}